\documentclass{article}

\usepackage{arxiv}

\usepackage[utf8]{inputenc}
\usepackage[T1]{fontenc}
\usepackage{hyperref}
\usepackage{url}
\usepackage{booktabs}
\usepackage{amsfonts}
\usepackage{amsmath}
\usepackage{float}
\usepackage{placeins}
\usepackage{nicefrac}
\usepackage{microtype}
\usepackage{graphicx}

\title{Seven Sources of Physical AI Capability Formation}

\author{%
  Gang Chen \\
  Zyllion Data Technology (Shanghai) Co.,Ltd. \\
  \texttt{cheng@zylliondata.com}
}
\begin{document}
\maketitle

\begin{abstract}
Capabilities relevant to Physical AI can arise from materially different formation histories. Existing bodies of literature are commonly organized by robot morphology, model architecture, learning algorithm, task family, or application domain, yet these taxonomies do not directly answer a more fundamental question: what gives rise to a particular capability? We define a capability-formation source as a factor that makes a material, identifiable contribution to the formation of a capability. Sources need not be entities or processes of the same kind and must be distinguished from system components and construction steps. Our evidence search is therefore restricted to candidate sources that can form or modify AI capabilities; knowledge from physics, chemistry, mathematics, or engineering is included only when it actually enters the capability-formation process.

We identify and calibrate seven non-exclusive sources: Recorded-Experience Formation (RE), Predictive-Modeling Formation (PM), Evaluative-Interaction Formation (EI), Surrogate-Environment Formation (SE), Mechanism-Grounded Formation (MG), Embodied-Coupling Formation (EC), and Evolution-Driven Formation (ED).

The study uses a reconstructive inductive design with theoretical saturation as its stopping criterion. We traced an existing research matrix back to primary studies, rebuilt and deduplicated the core literature, and established explicit coding rules. Before the challenge phase, we fixed the stopping rule and then conducted three rounds of maximum-difference and negative-case sampling. Candidate counterexamples included curriculum learning, self-supervision, social learning, active inference, open-ended learning, planning and search, neuro-symbolic architectures, digital twins, generative physical world models, developmental learning, and morphology-control co-design.

Within the literature scope, inclusion rules, and coding criteria fixed as of September 4, 2026, all 49 evidence records were explainable by the seven sources individually or in combination. None of the R1-R3 challenge rounds produced an irreducible eighth source, and R3 required neither a new core definition nor a substantive new boundary rule. We therefore claim theoretical saturation only within the stated scope; we do not claim logical completeness or exhaustive coverage of future possibilities.

The contribution is an auditable, comparable, and falsifiable vocabulary for distinguishing similarity in observed capability from similarity in the conditions under which that capability was formed. That distinction bears directly on explanation, transfer, replication, and the conditions required to reproduce capabilities, as well as on technological assets, dependencies, governance evidence, and geoeconomic foundations for capability formation. Our analysis concerns structure and traceability; it does not estimate category frequencies, effect sizes, investment returns, or national competitiveness.

\end{abstract}


\section*{1. Introduction}

One way to ask how intelligence is formed is to ask where capabilities come from: accumulated experience, prediction of the world, feedback from action, surrogate practice, laws and domain knowledge, body-environment coupling, or evolution across generations. These factors belong to different kinds of entities and processes, but they can be studied on a common relational basis: how each contributes to capability formation. Applied to Physical AI, the question becomes: through what distinguishable sources can the capabilities required by Physical AI be formed?

When the evidence for this study was first collected and reconstructed, Physical AI did not yet have a single cross-domain definition accepted across the relevant communities. The literature nevertheless provided a sufficiently coherent scope: the object of inquiry is AI, not physical or mechanical systems in general, and not all knowledge about the physical world. We therefore examine methods, training mechanisms, model structures, environments, mechanism knowledge, embodied relations, and evolutionary processes that can form or alter AI capabilities. Basic scientific or engineering knowledge counts as evidence only when it actually enters capability formation. A fixed experimental record may contribute through RE, a predictive structure through PM, and an explicit physical law or engineering mechanism through MG.

What a system can do and where that capability came from are different questions. Two robots may exhibit the same grasping, locomotion, or assembly capability, and two systems may both adapt successfully to the physical world, while relying on very different formative bases. One capability may be grounded in recorded experience, another in predictive structure, another in an evaluative feedback loop, a surrogate environment, explicit mechanisms, body-mediated coupling, or cross-generational evolution. These differences change what the capability depends on---data, models, simulators, mechanism knowledge, physical platforms, training infrastructure, or evolutionary search---and consequently change the conditions for transfer, replication, compliance review, substitution, and reproduction. Looking only at final performance, model scale, or whether a system is ``embodied'' can therefore collapse capabilities with different formative dependencies into the same label, or reduce a genuinely multi-source system to a single fashionable technology category.

Studying capability-formation sources matters at four levels. Scientifically, it allows apparently similar capabilities to be compared by their formative bases. From an engineering perspective, it exposes conditions governing transfer, replication, cross-embodiment reuse, and reproduction. For industry and investment, it broadens the relevant asset base beyond model weights to include data rights, predictive models, surrogate environments, mechanism knowledge, physical platforms, feedback loops, and evolutionary-search infrastructure. For governance and geoeconomics, it provides a way to identify critical dependencies, substitutability, evidentiary traceability, and locally accumulated foundations for capability formation. These are analytical dimensions implied by source differences, not empirically estimated frequencies, economic weights, or macro-level causal effects.

Accordingly, this paper asks: What identifiable sources can make a material contribution to the formation of specific capabilities relevant to Physical AI? And, as maximum-difference samples, boundary cases, and candidate counterexamples are progressively added, do any irreducible sources remain that cannot be explained by the existing categories?

We do not redefine Physical AI, treat the seven sources as components or necessary conditions of Physical AI, or prescribe a mandatory construction pipeline from scenarios to software and physical objects. This study is concerned solely with capability formation.

This study makes three contributions. First, it proposes and calibrates seven non-exclusive capability-formation sources, each with positive criteria, negative boundaries, and representative lineage anchors. Second, it establishes critical boundary pairs---including RE/EI, PM/SE, and EI/ED---to make adjacent sources consistently distinguishable. Third, through a traceable R0-R3 reconstruction and challenge process, it shows that additional theoretical sampling within the specified literature scope and cutoff date ceased to yield new irreducible sources, while R3 required neither a new core definition nor a substantive new boundary rule. On this basis, we conclude that the framework reached theoretical saturation within the stated scope.

\section*{2. Object of Analysis and Boundaries}

\subsection*{2.1 Object of analysis}

The unit of analysis is not an individual paper, company, robot model, or complete system. It is a specific target capability together with a clearly identifiable formative practice at a particular stage of its formation. A single system may draw on different sources at different formative stages, and a single paper may provide evidence for multiple sources. Multiple coding therefore denotes a combination of formative sources; it does not imply that the categories themselves overlap, nor that a system is composed of seven modules.

\subsection*{2.2 Decision criterion}

A candidate factor is treated as a potential formation source if the available evidence indicates that removing or replacing it would materially alter the acquisition, shaping, improvement, transfer, or extension of the target capability. The factor must also exert an independently identifiable material role at the analytical level adopted here. A factor that merely determines training order, deployment-time invocation, architectural hosting, an evaluation metric, or generic scheduling does not warrant a separate source category. ``Independently identifiable'' is not defined circularly as ``not fitting the existing seven''; it distinguishes formative contribution from adjacent analytical roles. These judgments reconstruct the methods and experimental settings reported in the literature; they do not constitute new intervention experiments and should not be read as strict experimental causal identification. Deployment-only invocation, inference, action selection, or safety filtering is not coded as a source unless it changes training or longer-term adaptation.

\subsection*{2.3 Exclusion rule}

Capability-formation source is a relational concept: it describes how a factor makes an identifiable, material contribution to acquiring, shaping, improving, transferring, or extending a particular capability. Different sources therefore need not be entities or processes of the same kind. A new algorithm name, system architecture, training schedule, deployment pattern, or organizational mechanism does not automatically constitute a new source. A factor becomes a candidate new source only when it contributes to capability formation in an independently identifiable way that cannot be explained by the existing sources individually or in combination.

\section*{3. Method}

\subsection*{3.1 Reconstructive inductive design}

The seven-source framework did not originate in a fully preregistered, de novo study. It emerged historically from earlier project materials. We therefore describe the method as reconstructive induction, with theoretical saturation as the stopping criterion, rather than retrospectively recasting it as a de novo grounded-theory study. We first traced methodological and system-level leads in the prior research matrix back to primary studies, removed duplicate records, marketing descriptions, and materials lacking methodological detail, and reconstructed the R0 core literature. Before the challenge phase began, we fixed the stopping rule and conducted three rounds of maximum-difference and negative-case sampling. R0 reconstructed the core records and yielded the initial seven-source framework; R1-R3 then subjected it to successive maximum-difference and negative-case challenges. Section 5.3-5.4 reports the additional evidence, candidate counterexamples, and boundary refinements introduced in each round.

\subsection*{3.2 Literature scope and sampling dimensions}

The public evidence base consists predominantly of English-language technical literature, with priority given to primary studies, formally peer-reviewed versions, and authoritative reviews. Frontier preprints from 2025-2026 are used only to stress-test boundaries, not to establish engineering maturity. The literature search cutoff is September 4, 2026. The final audit archive comprises 49 coded evidence records, 45 public sources, and six exclusion rules. A single public source may support more than one coded evidence record. Maximum-difference sampling spans offline, online, lifelong, and cross-generational timescales; real-world experience, recorded corpora, internal models, and external surrogate environments; evaluative signals such as reward, preference, self-supervised error, safety constraints, and fitness; and single-agent, multi-agent, collective, cross-embodiment, and morphology-control co-evolution settings.

\subsection*{3.3 Evidence inclusion and exclusion}

Included material must make it possible to identify what factor formed or changed a capability, rather than merely report improved performance. At least one source must be supported by the body of the paper, its methods, or its experimental setup; titles, abstracts, and common-sense inference alone are insufficient. Multiple coding is permitted for the same system. Records with insufficient evidence are left uncoded or coded conditionally.

Exclusions include claims of ``intelligence,'' ``embodiment,'' or ``generality'' without an account of capability formation; planning, search, or safety filtering used only at deployment time; mere inventories of hardware, sensors, or robot platforms; purely digital tasks with no formative relevance to the Physical AI lineage examined here; press releases and marketing materials lacking methodological detail; and duplicate versions of the same method.

\subsection*{3.4 Terminology calibration and coding rules}

Terminology backtracking confirmed that the earlier labels D/M/T/S/K/E/V can be mapped to Recorded-Experience Formation (RE), Predictive-Modeling Formation (PM), Evaluative-Interaction Formation (EI), Surrogate-Environment Formation (SE), Mechanism-Grounded Formation (MG), Embodied-Coupling Formation (EC), and Evolution-Driven Formation (ED). ``Formation'' is the shared semantic suffix and is not included in the two-letter abbreviations. This relabeling does not alter the substantive classification of the 49 records, the R0-R3 rounds, the evidence base, or the saturation result. MG does not mean merely ``mechanism-related'': it is coded only when an explicit mechanism materially participates in capability formation. Likewise, ED does not mean generic incremental improvement: it requires variation-selection-retention or inheritance across generations, populations, or evolutionary cycles to materially drive capability formation.

\subsection*{3.5 Stopping rule for theoretical saturation}

After R0 reconstruction and the first version of the coding rules were complete, the following stopping rule was fixed before the challenge phase: conduct three consecutive rounds of maximum-difference/negative-case sampling, each covering at least four candidate counterexample families; observe no new source irreducible to the seven categories in any round; require no substantive revision of any core category definition in the final two rounds; and ensure that every uncoded record can be explained as capability use, architecture, organization, an evaluation metric, or insufficient evidence. Sampling was to stop only if all four conditions were met.

\section*{4. Seven Sources of Capability Formation}

The seven categories are non-exclusive at the level of coding: for a given target capability, a single formation history or concrete formation pathway may involve multiple sources. This does not mean that the source definitions themselves are intentionally overlapping. The sources may be independently identifiable or combined within the same formation history or concrete formation pathway. They are not seven algorithm families, system components, or construction steps. Figure 1 expresses only the analytical relation between formation sources and a target capability.

The seven sources need not be equally mature in real-world implementation or system-level integration. Evidence for a source, physical realization of that source, and integration into a relatively complete system are three distinct questions. The discussion below reports the implementation status supported by currently available public evidence.

Naming note. Recorded-Experience Formation (RE), Predictive-Modeling Formation (PM), Evaluative-Interaction Formation (EI), Surrogate-Environment Formation (SE), Mechanism-Grounded Formation (MG), Embodied-Coupling Formation (EC), and Evolution-Driven Formation (ED) are analytical category names introduced here to provide a uniform vocabulary for the seven sources. The naming convention is ``core source concept + Formation,'' with Formation serving as a common semantic suffix rather than part of the abbreviation. Our claim regarding these names is limited to definitional clarity, discriminability, and internal consistency within this study. The technical phenomena, methods, and research lineages denoted by the names come from the existing literature.

For engineering context only, we note the current public scope of IEEE P4501 on Physical Artificial Intelligence in Manufacturing, which lists sensing, cognition, decision-making, and actuation among Physical AI components. Those components are not used as classification criteria for the seven sources, nor do we claim that the P4501 list constitutes an exhaustive system architecture \cite{ieee2026}.

\begin{figure}[H]
\centering
\includegraphics[width=\textwidth]{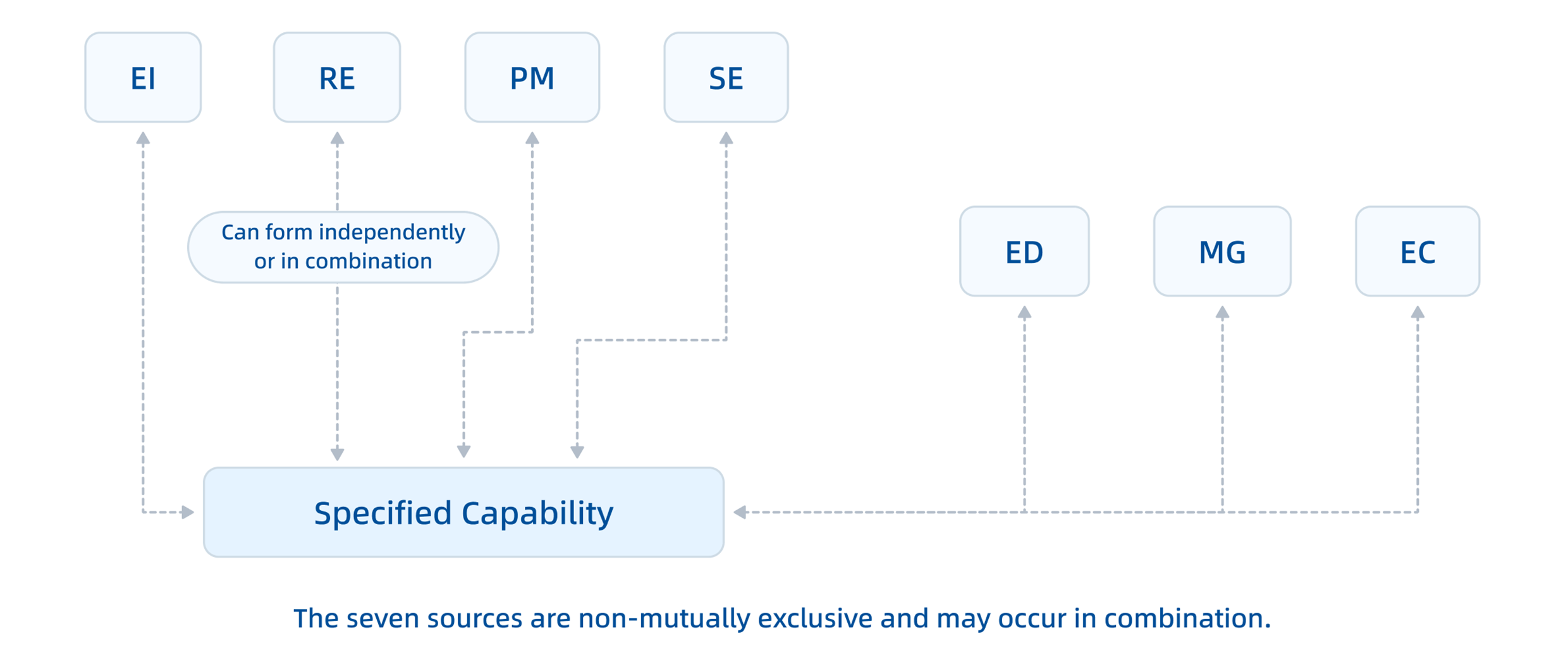}
\caption{\textit{Relationship Between the Seven Capability-Formation Sources and a Specified Capability.}}
\label{fig:fig1}
\end{figure}

\subsection*{4.1 Recorded-Experience Formation (RE)}

Recorded-Experience Formation: capability can be formed directly from fixed, recorded experience. RE denotes recorded and fixed observations, trajectories, text, images, preferences, demonstrations, or other pre-existing experience that materially participates in capability formation. The defining feature is not generic ``data-driven'' learning, but the role that recorded experience itself plays in forming the capability. Robot demonstrations, teleoperation trajectories, internet-scale vision-language corpora, and multi-robot datasets can all supply such material.

Positive criterion: fixed demonstration sets, robot trajectory repositories, web-scale pretraining corpora, or previously recorded state-action experience are used to form the capability. Negative boundary: merely stating that ``data were used'' is insufficient. If experience is generated online within the current action-consequence loop and fed back to update the capability, EI must also be considered.

Representative lineages include the survey of learning from demonstration by Argall et al.; ALOHA/Mobile ALOHA; Diffusion Policy; RT-2; Open X-Embodiment; OpenVLA; Octo; and PaLM-E \cite{argall2009,zhao2023,fu2024,chi2023,brohan2023,openx2023,kim2024,octo2024,driess2023}. These systems do not rely on RE alone, but their recorded demonstrations, trajectories, or pretraining corpora provide direct evidence for RE.

Real-system evidence: RT-2 forms capabilities from recorded robot trajectories and internet vision-language data and evaluates them on real robot-control tasks. This shows that RE can directly enter the formation of physical robot capabilities; it does not imply that RE by itself constitutes a complete Physical AI system \cite{brohan2023}.

\subsection*{4.2 Predictive-Modeling Formation (PM)}

Predictive-Modeling Formation: capability can be formed through prediction mediated by a world model. PM denotes learnable predictive structures---state transitions, future observations, action consequences, latent dynamics, or intervention-conditioned outcomes---that materially participate in capability formation. The key is not simply that a system ``has a model,'' but that predictive modeling contributes materially to training, policy formation, or capability improvement.

Positive criterion: a learned world model, dynamics model, or latent dynamics model participates in training, policy improvement, or the formation of planning capability. Negative boundary: a purely static classifier is not PM; nor is a model that is invoked only at deployment and does not change capability formation.

Representative lineages include Sutton's Dyna; Ha and Schmidhuber's World Models; PlaNet and Dreamer; DayDreamer; TD-MPC2; and V-JEPA-style predictive video representation learning \cite{sutton1991,ha2018,hafner2019,hafner2025,wu2022,hansen2024,bardes2024}. These systems are not PM-only systems.

Physical-system evidence: DayDreamer uses a learned world model for online learning on four real-robot tasks spanning quadruped locomotion, manipulation, and mobile navigation. It therefore provides direct evidence that PM can participate in physical robot capability formation, while the predictive model remains only one formative source within the complete system \cite{wu2022}.

\subsection*{4.3 Evaluative-Interaction Formation (EI)}

Evaluative-Interaction Formation: capability can be formed or refined through cycles of action and evaluative feedback. EI denotes a closed loop in which actions or candidate behaviors produce consequences that are evaluated through reward, preference, error, success/failure signals, prediction error, information gain, or other evaluative signals, and the resulting evaluation updates the capability. The term is deliberately broader than ``trial and error,'' because evaluation need not take the form of a single external reward signal.

Positive criterion: consequences or evaluation directly alter subsequent policy, representation, or adaptation. Negative boundary: interaction without evaluative updating is not EI. Fixed labels used as offline experience may instead fall under RE. Evolutionary fitness used only as a cross-generational selection signal is not automatically double-coded as EI.

Representative lineages include reinforcement learning and Dyna; Dreamer; reinforcement learning from human preferences; and active-inference research \cite{sutton1991,hafner2025,hansen2024,christiano2017,piolopez2016}. Active inference is coded as EI only when evaluative or error signals actually alter subsequent capability formation. What these approaches share is not a particular algorithm but an action/candidate behavior $\rightarrow$ consequence $\rightarrow$ evaluation $\rightarrow$ capability-update loop.

Real-interaction evidence: QT-Opt used more than 580,000 real robot grasp attempts to form a closed-loop visual grasping policy, showing that EI can form manipulation capability through a real action-consequence-evaluation-update cycle \cite{kalashnikov2018}.

\subsection*{4.4 Surrogate-Environment Formation (SE)}

Surrogate-Environment Formation: capability can be trained in a surrogate world. SE denotes capability formation within a simulated, modeled, or generative world that responds to actions and substitutes for the target physical environment as the site of formative interaction. The defining feature is the surrogate world as an experience-generating substrate, not merely having a predictor.

Positive criterion: the environment responds to actions and continuously generates interaction used for capability formation---for example, a physics simulator, interactive digital twin, model-internal training world, or action-controllable generative world. Negative boundary: static synthetic samples belong to RE; an internal predictor that is not instantiated as an interactive world remains PM.

Representative lineages include Shadow Hand, ANYmal learning control, massively parallel legged reinforcement learning, Walk These Ways, Humanoid-Gym, domain randomization, controller adaptation inside World Models, and emerging interactive generative worlds \cite{andrychowicz2020,hwangbo2019,rudin2022,margolis2023,gu2024,tobin2017,ha2018}. What they share is not a particular simulator, but a surrogate world that responds to actions and provides the interactions from which the capability is formed.

From surrogate environment to physical system: ANYmal locomotion policies were formed in physics simulation and transferred to a real quadruped, where they supported walking, running, and recovery from falls. This provides a clear sim-to-real example of SE \cite{hwangbo2019}.

\subsection*{4.5 Mechanism-Grounded Formation (MG)}

Mechanism-Grounded Formation: capability can be grounded in explicit physical mechanisms. MG denotes explicit, articulable physical laws, conservation relations, geometric structures, stability or safety constraints, dynamics, invariants, or other objective mechanisms that materially participate in capability formation through priors, model structure, loss functions, feasible sets, policy-update constraints, or direct computational definitions. ``Grounded'' is intended to cover mechanisms entering as constraints, priors, structures, or direct definitions.

Positive criterion: removing the mechanism knowledge would materially change the learnable space, training objective, policy formation, or capability boundary. Negative boundary: generic regularization, merely being ``physically plausible,'' simply using a physics engine, post-hoc mechanistic explanation, or deployment-only safety filtering does not automatically constitute MG.

Representative lineages include physics-informed neural networks (PINNs), Hamiltonian neural networks, and training-time safe reinforcement learning with control barrier functions \cite{raissi2019,greydanus2019,cheng2019}. What they share is not simply the use of physics, but the fact that an explicit mechanism enters capability formation as a prior, structure, loss, constraint, or direct computational definition.

Physical experiments: work embedding physical structure into robot-learning models and validating it on soft-robot motion prediction and Franka Emika Panda trajectory tracking supports the participation of MG in physical modeling and control capability formation. The current evidence primarily concerns local capabilities and control loops rather than complete Physical AI systems \cite{liu2024}.

\subsection*{4.6 Embodied-Coupling Formation (EC)}

Embodied-Coupling Formation: capability can be shaped through body-environment interaction. EC denotes a material role in capability formation for body morphology, material properties, sensorimotor loops, contact dynamics, or body-environment coupling itself. The body is not a passive carrier to which an already formed capability is later attached; it participates in the formation process.

Positive criterion: changing morphology, material properties, sensorimotor relations, or coupling dynamics materially changes what capability can be formed, how it is formed, or the cost of forming it. Negative boundary: merely having a robot body, using a URDF, or ultimately deploying on a physical platform is insufficient. The shaping effect can also occur during simulation-based training.

Evidence for EC must be separated into two levels. Passive dynamic walking and soft-body morphological computation provide strong evidence that bodily dynamics can constitute behavioral capability or computational resources. This establishes the mechanistic basis for EC and demonstrates that the body can materially participate in intelligent processes. But ``the body contributes to behavior/computation'' and ``the body contributes to capability formation'' are not the same claim. More direct evidence for EC comes from studies showing that changes in the body alter how capabilities are learned or how the policy space is shaped---for example, morphology-control co-design and comparative experiments in which different morphologies systematically change policy learning. We therefore present these two evidentiary levels separately rather than treating the first as equivalent to the second.

Representative lineages include McGeer's passive dynamic walking, the morphological-computation tradition, ANYmal, Shadow Hand, and morphology-control co-design \cite{mcgeer1990,muller2017,hwangbo2019,andrychowicz2020,cheney2018}. The relevant distinction is not embodiment per se, but whether morphology, material, or coupling dynamics actually shape capability formation.

Physical coupling evidence: experiments with soft silicone arms show that bodily dynamics can themselves serve as resources for real-time computation and closed-loop control, supporting the physical basis of EC at the mechanism level. Morphology-control co-design goes further by showing that bodily changes can alter the path and cost of policy learning, providing more direct evidence of EC at the level of capability formation. These studies concern how body-environment dynamics participate in capability formation; they do not by themselves establish complete system-level Physical AI integration \cite{nakajima2015,cheney2018}.

\subsection*{4.7 Evolution-Driven Formation (ED)}

Evolution-Driven Formation: capability can be formed through selection and elimination across generations. ED specifically denotes capability formation at cross-generational, population, or evolutionary-cycle timescales, where variation, selection, and inheritance/retention materially drive the formation of controllers, morphologies, task ecologies, or capability distributions. ``Driven'' emphasizes that the evolutionary mechanism itself is formative rather than a loose synonym for a system becoming better over time.

Positive criterion: an identifiable variation, selection, and retention or inheritance process changes capability across generations, populations, or evolutionary cycles. Negative boundary: within-lifetime gradient updates, ordinary reinforcement learning, self-improvement, developmental learning, or iterative optimization without retention/selection do not constitute ED.

Representative lineages include evolutionary robotics, embodied evolution, MAP-Elites and quality-diversity methods, POET, and morphology-control co-optimization \cite{nolfi2000,bredeche2018,mouret2015,wang2019,cheney2018}. What they share is variation, selection, and retention or inheritance across generations, populations, or evolutionary cycles, not generic continuous improvement.

Physical evolutionary evidence: real-world evolution experiments have jointly searched the morphology and control of mechanically reconfigurable quadruped robots using physical evaluation, while also noting that much evolutionary-robotics work relies on physics simulators. ED therefore has direct physical experimental support, although current representative evidence is concentrated in experimental evolutionary-robotics systems \cite{nygaard2021}.

\FloatBarrier
\subsection*{4.8 Definition-Boundary-Lineage Matrix}

\begin{table}[htbp]
\centering
\footnotesize
\setlength{\tabcolsep}{4pt}
\begin{tabular}{@{}p{0.8cm}p{4.3cm}p{3.3cm}p{3.5cm}p{2.6cm}@{}}
\toprule
\textbf{Source} & \textbf{Core definition} & \textbf{Positive criterion} & \textbf{Negative boundary} & \textbf{Lineage anchors} \\
\midrule
RE & Pre-existing recorded experience materially participates in capability formation. & Fixed demonstrations, trajectories, corpora. & Generic ``data-driven'' claims; online feedback requires EI check. & Learning from demonstration; ALOHA; RT-2 \\
PM & A learnable predictive structure materially participates in formation. & World models; latent dynamics. & Static classification; deployment-only model invocation. & Dyna; Dreamer; TD-MPC2 \\
EI & Consequences or an evaluative loop change capability. & Reward, preference, error, success signals. & Interaction without evaluative updating. & Reinforcement learning; preference learning \\
SE & An action-responsive surrogate world hosts formative interaction. & Simulators; interactive twins. & Static synthetic samples; internal prediction only. & Shadow Hand; ANYmal \\
MG & Explicit mechanisms materially enter capability formation. & PDE residuals; Hamiltonians; barrier functions. & Generic regularization; runtime-only constraints. & PINNs; HNNs; safe RL \\
EC & Body, material, or coupling materially shapes formation. & Morphology; contact dynamics. & Having a body/URDF/deployment alone. & Morphology-control co-design; passive dynamics \\
ED & Cross-generational/population variation-selection-retention drives formation. & Evolutionary cycles; archives. & Ordinary RL; self-improvement; iterative optimization. & Evolutionary robotics; QD; POET \\
\bottomrule
\end{tabular}
\end{table}

\FloatBarrier
\section*{5. Results: Distinction, Combination, and Theoretical Saturation}

\subsection*{5.1 Critical adjacent boundaries}

A seven-source framework is useful only if adjacent categories remain distinguishable in difficult cases. The most consequential boundary pairs are summarized below.

\begin{table}[htbp]
\centering
\footnotesize
\setlength{\tabcolsep}{4pt}
\begin{tabular}{@{}p{2.2cm}p{7.4cm}p{5.2cm}@{}}
\toprule
\textbf{Boundary pair} & \textbf{Fixed criterion} & \textbf{Misreading prevented} \\
\midrule
RE / EI & Experience fixed before updating $\rightarrow$ RE; action consequence/evaluation immediately changes subsequent capability $\rightarrow$ EI. & Both may occur at different stages of one project. \\
PM / SE & Predictive structure $\rightarrow$ PM; instantiated as an action-responsive surrogate world that hosts training $\rightarrow$ SE. & Prevents all predictive rollouts from being labeled SE. \\
EI / ED & Within-lifetime evaluative updating $\rightarrow$ EI; cross-generational or population-level variation, selection, and retention $\rightarrow$ ED. & Evolutionary fitness used only for selection is not double-coded. \\
RE / SE (generated data) & Fixed generated artifacts/stored samples $\rightarrow$ RE; continuously action-responsive training world $\rightarrow$ SE. & The generator's predictive structure may additionally constitute PM. \\
\bottomrule
\end{tabular}
\end{table}

\begin{figure}[!htbp]
\centering
\includegraphics[width=\textwidth]{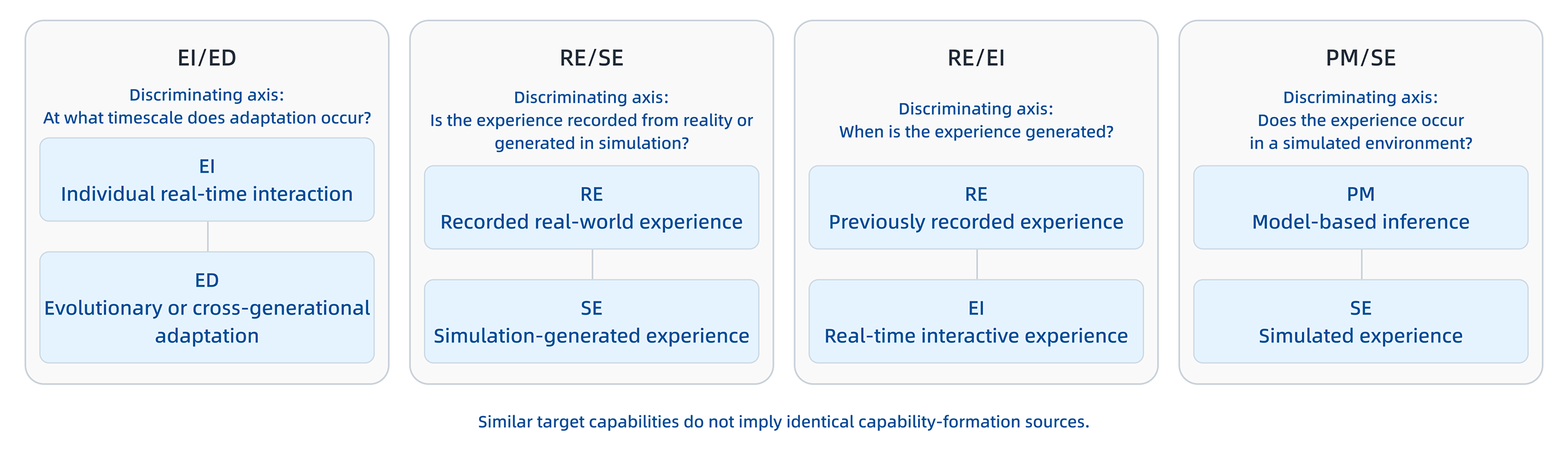}
\caption{\textit{Key Adjacent Boundaries Among the Seven Capability-Formation Sources}}
\label{fig:fig2}
\end{figure}

\subsection*{5.2 Combination without collapse}

Combining sources does not make the distinctions between them disappear. PlaNet and Dreamer, for example, can combine PM, SE, and EI. Shadow Hand and ANYmal can combine SE, EI, and EC. POET can combine ED, SE, and EI. Physics-informed reinforcement learning can combine MG and EI and, where appropriate, PM. The framework instead makes clear that a system's formation history can superimpose several independently meaningful contributions, each imposing different requirements for transfer, replication, and substitution.

\subsection*{5.3 Negative-case tests for an ``eighth source''}

\begin{table}[htbp]
\centering
\footnotesize
\setlength{\tabcolsep}{4pt}
\begin{tabular}{@{}p{3.4cm}p{4.0cm}p{7.8cm}@{}}
\toprule
\textbf{Candidate concept} & \textbf{Actual analytical level} & \textbf{Reduction under the seven-source framework} \\
\midrule
Curriculum learning & Training organization/order & Reorders RE/EI/SE; does not introduce a new formation source \cite{narvekar2020}. \\
Self-supervised learning & Construction of supervision signals & Fixed corpus $\rightarrow$ RE; predictive task $\rightarrow$ PM. \\
Transfer learning / foundation models & Knowledge-transfer mechanism & Transferred recorded experience $\rightarrow$ RE; architecture itself is not a source. \\
Social / multi-agent learning & Provider and interaction structure of experience & Observation $\rightarrow$ RE; feedback $\rightarrow$ EI; body-environment interaction $\rightarrow$ EC; population inheritance $\rightarrow$ ED. \\
Active inference & Unified theory/control scheme & Generative model $\rightarrow$ PM + error loop $\rightarrow$ EI; EC/MG where warranted. \\
Open-ended learning / intrinsic motivation & Regime for generating goals and tasks & EI + SE + ED; novelty is an evaluative quantity. \\
Planning / search / MPC & Capability-use or optimization operator & Not coded if runtime-only; PM/EI when it participates in formation. \\
Neuro-symbolic / modular systems & Representation and construction layer & May host RE/PM/MG; architecture itself is not a source. \\
Digital twins & System form & Prediction $\rightarrow$ PM; interactive surrogate $\rightarrow$ SE; mechanistic consistency $\rightarrow$ MG. \\
Generated video / synthetic data & Data-production mechanism & Fixed output $\rightarrow$ RE; interactive environment $\rightarrow$ SE; predictor $\rightarrow$ PM. \\
Causal representation learning & Model structure / identification objective & Prediction of intervention outcomes $\rightarrow$ PM; explicit mechanism grounding $\rightarrow$ MG. \\
Developmental learning & Temporal organization & Within-lifetime RE/EI/EC; not ED without cross-generational selection. \\
Co-design / automated machine design & Expanded search object & Body shaping $\rightarrow$ EC; cross-generational search $\rightarrow$ ED; mechanistic restrictions $\rightarrow$ MG where explicit. \\
\bottomrule
\end{tabular}
\end{table}

Social/collective formation and open-ended task generation remain important candidates for continued scrutiny. The former changes the inter-agent structure through which experience is supplied; the latter continually creates goals, environments, and evaluative conditions. In the present evidence, however, the former remains reducible to RE/EI/EC/ED and the latter to PM/SE/EI/ED. Neither currently warrants an eighth source at the same level of analysis.

\subsection*{5.4 R0-R3 saturation trajectory}

R0 reconstructed 28 core evidence records, yielding the initial seven-source framework and establishing early boundaries including PM/SE and EI/ED. R1, R2, and R3 each added seven maximum-difference/negative-case records, yet none produced a new irreducible source. R2 fixed the rules that deployment-only use is not coded and that fixed generated artifacts fall under RE whereas interactive generated worlds fall under SE. When R3 confronted physical world models, interactive generative worlds, physics-informed reinforcement learning, vision-language-action models, developmental learning, and morphology-control co-design, it produced neither a new source nor a need to revise the seven core definitions or add a substantive boundary rule. R2 was therefore the final round of substantive boundary refinement, and R3 served as a subsequent stability confirmation.

\begin{figure}[!htbp]
\centering
\includegraphics[width=0.6\textwidth]{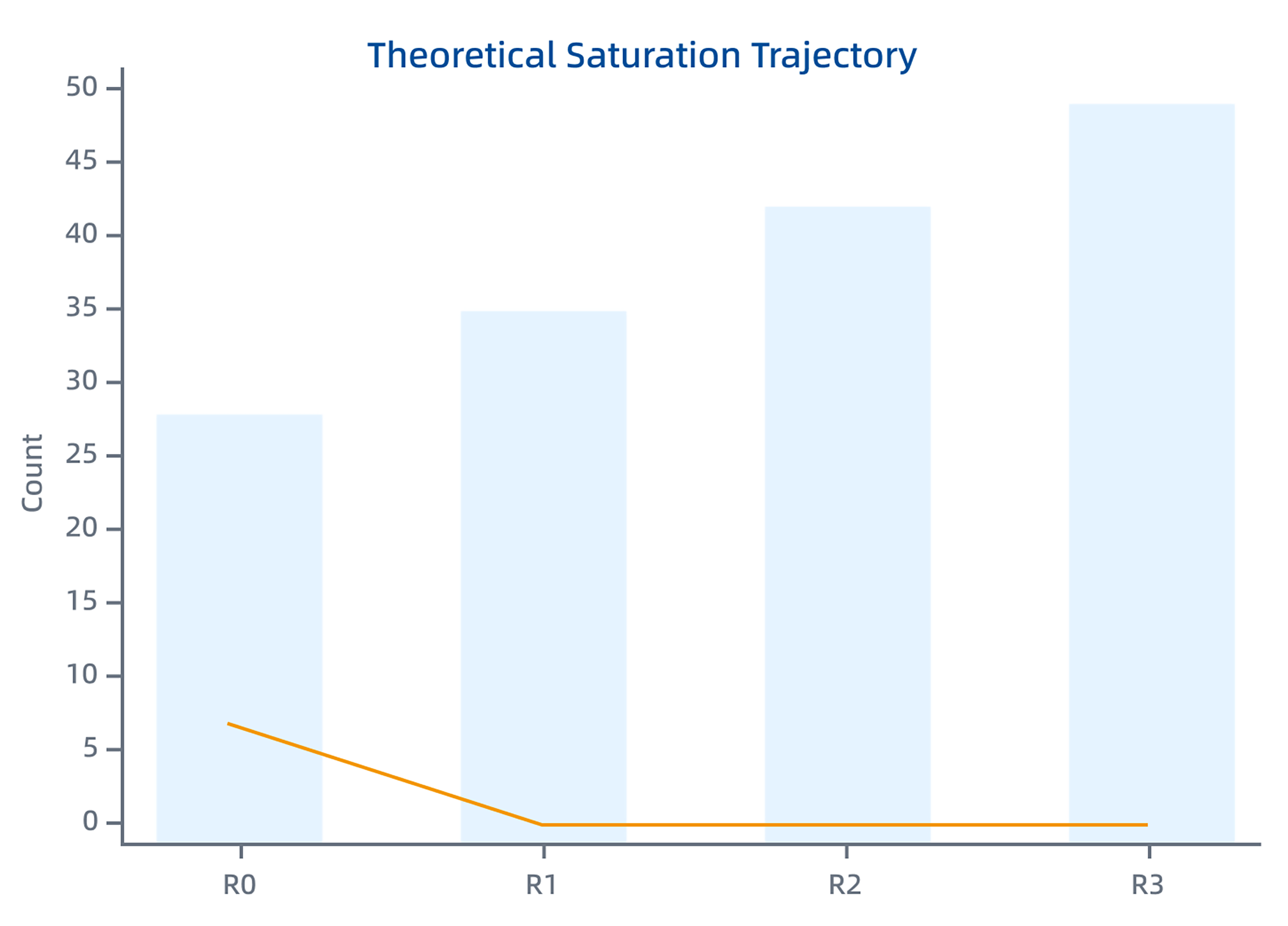}
\caption{\textit{Theoretical Saturation Trajectory of the Seven Capability-Formation Sources}}
\label{fig:fig3}
{\footnotesize\textit{Bars: Cumulative evidence records = 28 $\rightarrow$ 35 $\rightarrow$ 42 $\rightarrow$ 49}\par}
{\footnotesize\textit{Line: Irreducible new sources = 7 $\rightarrow$ 0 $\rightarrow$ 0 $\rightarrow$ 0}\par}
{\footnotesize\textit{R2 marks the last substantive boundary refinement; R3 serves as a stability confirmation.}\par}
\end{figure}

We therefore stopped expanding the set of categories and report scope-qualified theoretical saturation. The scope is jointly defined by the specified literature domain, inclusion and exclusion rules, coding criteria, and the September 4, 2026 search cutoff. Saturation refers only to whether continued sampling produces additional irreducible capability-formation source categories. It does not imply exhaustion of within-category subtypes, combinations or their frequencies, causal effect sizes, industry distributions, or future technologies, and it is not a claim of logical completeness.

\subsection*{5.5 Robustness checks}

To reduce the risk that high-level technology labels would induce the coding, we applied several robustness checks addressing label ablation, temporal layer, where the relevant process occurs in the system, where knowledge enters capability formation, negative cases, and source combinations. Vision-language-action models, world models, active inference, and digital twins were coded from evidence about capability formation after their labels were stripped away. We distinguished experience fixed before training, online within-lifetime updating, deployment-time use, and cross-generational evolution; separated predictive structures (PM), instantiated action-responsive surrogate environments (SE), and bodily coupling (EC); and coded MG only when an explicit mechanism entered capability formation.

\section*{6. Research Significance and Application Value}

We do not interpret the number seven as a score of performance, maturity, valuation, or national competitiveness. This section instead identifies audit questions and decision dimensions that follow from distinguishing formation sources; it does not report empirically measured source frequencies, effect sizes, economic value, investment returns, or national competitiveness. In practice, the framework shifts the question from ``Which fashionable technology does this system use?'' to asking ``On which formative sources does this capability depend, and can those sources be accessed, transferred, reproduced, or substituted?''

\subsection*{6.1 Investment perspective}

For investors, the framework turns the vague notion of a ``capability moat'' into a set of traceable questions. Does the core capability depend primarily on irreplaceable recorded experience or on a replicable predictive model? Are simulation infrastructure, mechanism knowledge bases, specialized physical platforms, or cross-generational evolutionary experiments critical formative assets? When a company claims to possess ``robot intelligence,'' the framework prompts further questions: Does it control the relevant data rights, simulators, mechanism expertise, hardware embodiments, and training loops? Where, exactly, does the cost to a competitor of reproducing an equivalent capability arise?

\subsection*{6.2 AI-company perspective}

For AI companies, the framework separates observed model capability from the conditions under which that capability was formed. Heavy dependence on RE may make cross-scenario expansion primarily a problem of experience coverage and data rights. Dependence on PM or SE shifts attention toward world-model quality and interactive surrogate environments. Dependence on MG makes the acquisition, formalization, and governance of mechanism knowledge part of the basis of capability formation. Dependence on EC means that copying software weights alone may not reproduce the capability. Dependence on ED makes evaluation cycles, population scale, retention mechanisms, and search infrastructure part of the cost of capability formation.

\subsection*{6.3 Industrial-enterprise perspective}

Industrial enterprises evaluating or deploying Physical AI should not assess only the final demonstration. They should ask under what conditions the capability was formed. Systems that depend on extensive proprietary site experience may incur transfer costs primarily through recollecting and cleaning data. Systems dependent on SE require validation of the surrogate environment against real operating conditions. Systems dependent on MG require audits of whether the mechanism is correct, maintainable, and genuinely operative during formation. Systems dependent on EC require assessment of whether changes in bodies, fixtures, materials, or contact conditions will invalidate an existing capability. The framework thus supplies an audit vocabulary for transfer, maintenance, and retraining.

\subsection*{6.4 Public-governance perspective}

For regulators and other public-sector decision makers, the seven-source framework helps separate how an AI capability was formed from runtime safety controls. A system may be subject to mechanism-based constraints at runtime without having formed its capability through MG; conversely, it may undergo extensive SE-based training while remaining exposed to uncovered conditions after real-world deployment. Regulatory assessment can therefore examine formative evidence, deployment constraints, reproducible experiments, and the provenance of data and simulators separately, rather than treating ``uses physical constraints'' or ``trained in simulation'' as a single safety label.

\subsection*{6.5 Geoeconomic perspective}

From a geoeconomic perspective, the framework is better suited to identifying the foundations on which capabilities are formed than ranking countries. Economies may differ in their accumulated data scale and data rights for RE; foundation models and compute for PM; simulation and digital-twin infrastructure for SE; industrial mechanism knowledge and engineering talent for MG; robot embodiments and supply chains for EC; and large-scale experimentation and open-ended search capacity for ED. The more useful policy question is therefore not ``Who possesses Physical AI?'' but which formative sources underpin strategically important capabilities, which can be imported, copied, or transferred, and which require long-term local accumulation. We do not use the framework to rank countries, nor do we treat these implications as empirically validated macroeconomic conclusions.

\section*{7. Limitations, Falsification, and Reopening Procedure}

First, the search is not an enumerable census of the entire internet. Results are constrained by language, database indexing, access, and the search cutoff, and the evidence base is predominantly English-language technical literature. Second, the study did not use independent dual coders and therefore does not report Cohen's $\kappa$ or another inter-coder reliability coefficient. Future validation should include independent coding of a substantial subset of the records by researchers who did not participate in developing the framework. Third, some 2025-2026 materials are preprints used only for boundary stress testing. Fourth, evidence records are organized around representative methods, so record counts cannot be interpreted as category importance, prevalence, or causal effect size.

The framework is explicitly falsifiable. R4 should be opened if future evidence shows that a capability depends materially on a formative contribution that cannot be reduced to any of the seven existing sources, individually or in combination. ``Irreducible'' does not mean a new name, algorithm, system-level implementation, or technology school. It means that after the current coding rules, seven definitions, and boundary rules are frozen, the candidate factor still cannot be fully explained by the existing sources individually or in combination and requires a new, independently identifiable capability-formation source. Social or collective processes should also trigger reconsideration if they exhibit a formative contribution irreducible to information, evaluation, coupling, or selection. Material self-organization, physical self-construction, and quantum, chemical, or biohybrid systems may likewise require extension if they introduce genuinely new timescales or units of inheritance.

Any reopening should begin by freezing the current version and its search cutoff. Reviewers should receive the coding rules but not the prior judgments and independently code the new records. If disagreements reduce to insufficient evidence, temporal-layer confusion, or an existing boundary-pair issue, the audit record should be updated without adding a category. A new category is warranted only when a genuinely new irreducible capability-formation source appears.

\section*{8. Conclusion}

This paper asks not what Physical AI is made of, but where particular capabilities relevant to Physical AI come from. Based on reconstructive induction, 49 evidence records, 45 public sources, and three rounds of maximum-difference and negative-case challenges, we retain seven non-exclusive sources: Recorded-Experience Formation (RE), Predictive-Modeling Formation (PM), Evaluative-Interaction Formation (EI), Surrogate-Environment Formation (SE), Mechanism-Grounded Formation (MG), Embodied-Coupling Formation (EC), and Evolution-Driven Formation (ED).

The framework is not intended to force whole systems into rigid categories. Its purpose is to expose different formative dependencies that can sit behind similar technical labels. Fixed experience is not the same as online evaluation; predictive structure is not the same as a surrogate world; within-lifetime adaptation is not the same as cross-generational evolution; explicit mechanism grounding is not equivalent to vaguely ``using physics''; and having a body is not equivalent to the body actually shaping capability formation. A system may combine several sources, but combination does not erase these distinctions.

Under the specified literature scope, inclusion rules, coding criteria, R0-R3 stopping rule, and September 4, 2026 search cutoff, three successive challenge rounds produced no irreducible eighth source, and R3 introduced neither a new core definition nor a substantive new boundary rule. We therefore report scope-qualified theoretical saturation with respect only to the emergence of additional irreducible capability-formation source categories. This does not mean that the seven sources are logically complete for all time, nor that their internal subtypes, combinations, effect sizes, industry distributions, or future technological manifestations have been exhausted. The result is instead an auditable, comparable, and falsifiable framework that can be overturned or extended by new evidence.

\section*{Supplementary Audit File}

The complete set of 49 evidence records, 45 public sources, the R0-R3 saturation log, exclusion rules, candidate counterexamples, and prospective R4 triggers is preserved in the accompanying auditable saturation-induction research archive. The main paper reports only the definitions, boundaries, representative lineages, and saturation results necessary to support the research question.

\end{document}